\pdfoutput=1
\def\year{2019}\relax
\documentclass[letterpaper]{article} 
\usepackage{aaai19}   
\usepackage{times}    
\usepackage{helvet}   
\usepackage{courier}  
\usepackage{url}  
\usepackage{graphicx}  
\usepackage{graphicx}
\usepackage{amsmath}
\usepackage{amssymb}
\usepackage{epsfig}
\usepackage{array}
\usepackage{multirow}
\usepackage{rotating}
\usepackage{subfig}
\usepackage{enumitem}
\usepackage{wrapfig}

\begin{document}
%
\title{Learning Embeddings for Product Visual Search with \\ Triplet Loss and Online Sampling}

\author{
   Eric Dodds, Huy Nguyen, Simao Herdade, Jack Culpepper, Andrew Kae, Pierre Garrigues \\
   Yahoo Research \\
   \texttt{\{eric.mcvoy.dodds, huyng, sherdade, jackcul, andrewkae, garp\}@oath.com} \\
}

\maketitle




\begin{abstract}
 In this paper, we propose learning an embedding function for content-based image retrieval within the e-commerce domain using the triplet loss and an online sampling method that constructs triplets from within a minibatch. We compare our method to several strong baselines as well as recent works on the DeepFashion and Stanford Online Product datasets.  Our approach significantly outperforms the state-of-the-art on the DeepFashion dataset. With a modification to favor sampling minibatches from a single product category, the same approach demonstrates competitive results when compared to the state-of-the-art for the Stanford Online Products dataset.

\end{abstract}

\section{Introduction}
\label{sec:intro}

Visual search is an increasingly popular tool that enables users to quickly find similar products from large online product catalogs. At its core, a visual search system embeds both query and catalog images into a common feature space and, at search-time, uses this feature space to retrieve a query image's k-nearest neighbors in the catalog. Whether such a system is able to surface relevant results for any given query depends heavily upon the quality of the embedding function used to represent both the query and catalog images.

Several approaches have been proposed in the shopping domain to learn a retrieval model from labeled pairs of images of the same item ~\cite{Huang2015,Kiapour2015,Liu2016,Gajic2018}. In these methods, a model is trained to embed images such that ``positive" pairs of images with the same items are represented with features closer to each other than ``negative" pairs of images containing different items.

In this work, we develop an image retrieval model using a deep convolutional neural network trained with a triplet loss. We show that a variant of ``batch-hard" triplet sampling~\cite{Hermans2017} allows this model to significantly improve on the state of the art for the DeepFashion consumer-to-shop clothes retrieval dataset~\cite{Liu2016}. We  demonstrate via ablation that the model improves further if we include all matched pairs during training, rather than only considering pairs that cross the query and catalog image sets. In addition, we evaluate our method on the Stanford Online Product dataset~\cite{Song2016} and show that we can obtain competitive results with current state-of-the-art methods by forming minibatches wherein all examples come from the same class for a portion of training time. We further analyze through exploratory analysis when it is advantageous to use such a sampling approach.

\begin{figure}
    \centering
    \includegraphics[width=0.45\textwidth]{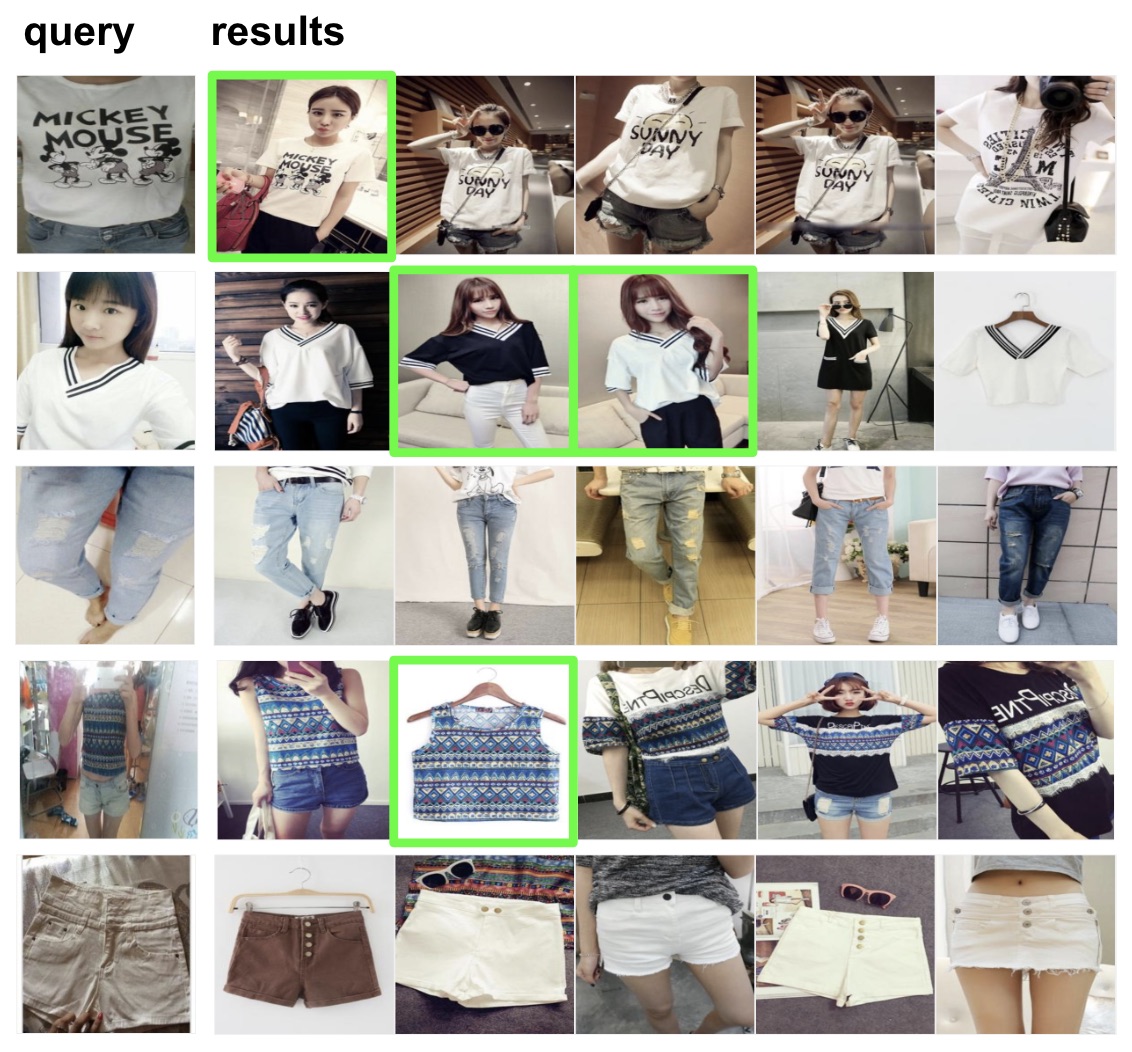}
    \caption{Top-5 image retrieval results using our proposed method for a few randomly sampled queries on the DeepFashion test dataset~\cite{Liu2016}. Correctly retrieved items are highlighted in green.}
    \label{fig:qualitative}
\end{figure}

\section{Related work}

\textbf{Content-based Image Retrieval} has been a well studied field of research. Babenko et al.~\shortcite{Babenko2014} show that features extracted from the penultimate layers of convolutional neural networks previously trained on ImageNet~\cite{ILSVRC15} outperform handcrafted features for retrieving semantically similar images when using euclidean distance as the distance metric. They further show that it is possible to train these same neural networks using classification data to improve retrieval performance for a given domain of interest.

\textbf{Deep Metric Learning}: While training to perform surrogate tasks such as classification has been shown to be an effective way to learn features for image retrieval~\cite{Babenko2014}, Deep Metric Learning (DML) methods aim to directly optimize a neural network to project data such that pairs of examples that are labeled ``similar" are represented with features that are close together in metric space and pairs of examples that are labeled ``disimilar" to be far apart in metric space~\cite{chopra2005learning}.  These approaches have been used to learn embeddings in a wide variety of tasks such as Face Verification~\cite{chopra2005learning}, Person Re-identification~\cite{Hermans2017}, and in our case, visual product search and fashion retrieval \cite{Kiapour2015,Liu2016,Huang2015}. Our work adopts a deep metric learning based approach to embed both query and catalog images into a metric space for image retrieval using the Triplet loss~\cite{Schroff2015} for training.



\section{Model and Sampling Approach}
\label{sec:training}
For a query image $x$, we retrieve the catalog images $y$ that are most similar to $x$ by the cosine similarity of their embeddings:
\begin{equation}
s(x, y) = \frac{f(x) \cdot f(y)}{||f(x)|| ||f(y)||}.
\end{equation} In our model, the embedding function $f$ is a ResNet-50-v2~\cite{He2016} neural network that was pre-trained on ImageNet~\cite{ILSVRC15}, with the classification layer removed.
The network is fine-tuned on the DeepFashion training set using the standard triplet loss~\cite{Schroff2015}: \begin{equation}\mathcal{L}_\text{triplet} = \frac{1}{|T|}\sum_{(a, p, n) \in T}  [s(a,n) - s(a,p) + m]_+.
\end{equation} Here $T$ is the set of (anchor, positive, negative) triplets and $m$ is a fixed margin hyperparameter. We set $m$ to 0.1 for the results shown in this paper. The choice of base network architecture and loss function follow \cite{Gajic2018}.

The key to the success of our approach is in sampling effectively from the set of triplets $T$. The number of triplets is cubic in the size of the dataset and therefore impractical to cover entirely during training. We observe empirically that uniform sampling leads to poor performance, worse than the pretrained baseline for the hyperparameter settings we tried. Various sampling methods have been proposed to combat this issue~\cite{Schroff2015,Huang2015,Gajic2018}. We adapt the ``batch-hard" sampling technique proposed in \cite{Hermans2017} in the person re-ID setting. We construct a minibatch by first drawing ``anchor" images uniformly at random. A ``positive" image $p_i$ for each anchor $a_i$ is chosen at random from the set of images that contain the same item as the anchor. For each pair $(a_i, p_i)$ in the minibatch, we form one triplet by setting the negative $n_i$ to be the ``positive" $p_{j \neq i}$ from another pair within the batch, that is closest to the anchor $a_i$:
\begin{equation}
    n_i = p_j \text{~~~where~~~} j = \underset{k}{\text{argmax~}} s(a_i,p_k).
\end{equation} The loss and its gradient for the minibatch are evaluated only on these selected triplets. The selection process can also be extended to the anchors $a_{j \neq i}$ when appropriate for the dataset. Our method differs from ``batch-hard" in that we sample positives uniformly, with only the negatives selected to be hard. The original batch-hard method instead created minibatches with several images of each item in the minibatch and then selected positives as well as negatives to find the hardest triplets.


\begin{figure}
    \centering
    \includegraphics[width=0.35\textwidth]{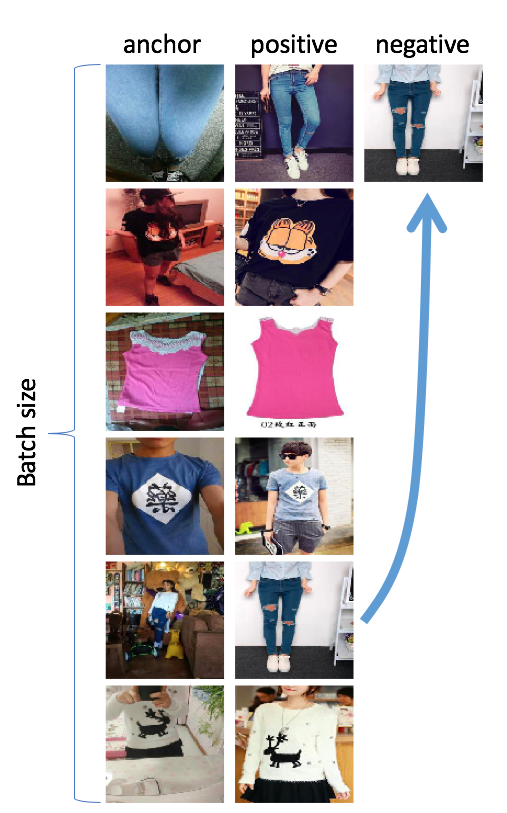}
    \caption{Illustration of our triplet sampling approach. Each matching pair forms a triplet with one non-matching example, chosen as the most similar to the anchor in the minibatch.}
    \label{fig:sampling}
\end{figure}

\section{Experimental Evaluation}
We evaluated our methods on two publicly available image retrieval datasets with images from e-commerce websites.
\label{sec:experiments}
\subsection{DeepFashion}

DeepFashion Consumer-to-Shop Clothes Retrieval~\cite{Liu2016} is a popular dataset for evaluating the image retrieval task in the fashion domain. The Consumer-to-Shop Clothes Retrieval benchmark within this dataset contains 251,361 consumer-to-shop image pairs from online retailer \textit{Mogujie}. Each image has a bounding box for one of 33,881 items, with each item belonging to one of 23 high-level clothing categories. The training/validation/test splits of the dataset consist of non-overlapping subsets of the clothing items.
The photos exhibit a wide diversity in appearance. Some photos come from an ``in-the-wild" setting while others come from online catalogs where the clothing items often have pristine white backgrounds. The consumer images are all ``in-the-wild" while the shop images tend to provide more ideal views of the labeled clothing item. We use the item labels to form the anchor-positive pairs for training our image retrieval model as described in Model and Sampling Approach.


\begin{figure}
    \centering
    \includegraphics[width=0.45\textwidth]{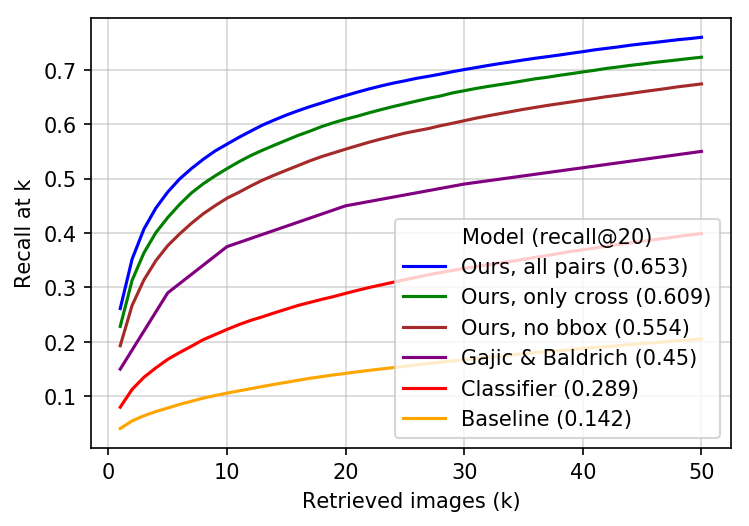}
    \caption{Recall at k of our model on the DeepFashion dataset compared to several baseline methods and the previous state-of-the-art from \cite{Gajic2018}.}
    \label{fig:ret@k}
\end{figure}

In figure \ref{fig:ret@k}, we use the recall at k metric~\cite{Kiapour2015} to compare our model to several other methods commonly used to learn image feature embeddings for retrieval. All models shown in this figure use the same Resnet50-v2 base architecture. Our method (blue) outperforms generic embeddings extracted from a network that had been pre-trained on ImageNet~\cite{ILSVRC15} (orange). It also outperforms embeddings extracted from a network trained to predict one of the 23 DeepFashion classes using a softmax cross-entropy loss (red).
The previous state-of-the-art method proposed by Gajic and Baldrich~\cite{Gajic2018} (purple) uses a triplet loss but relies on an offline hard triplet mining approach, unlike our online batch-hard sampling. We show that our approach can retrieve an exact item match in the first 20 retrieved images over 65\% of the time as compared to Gajic and Baldrich's 45\% retrieval accuracy. Our approach represents a 44\% relative improvement over the previous state-of-the-art result.

Although the model is tested by querying with ``consumer" images and retrieving from a separate set of ``shop" images, our model improves if we ignore the distinction between these domains during training. The retrieval performance of our model trained only using cross-domain pairs is shown in green in figure \ref{fig:ret@k} for comparison.

DeepFashion provides a bounding box for each image, indicating the item the image is labeled as containing. We cropped to these bounding boxes for our primary experiments, but we also trained and evaluated the model using the whole images, ignoring the provided bounding boxes. The results of this experiment are shown in brown in figure \ref{fig:ret@k}.

\subsection{Qualitative Results}

In figure \ref{fig:qualitative}, we show some retrieval results from the DeepFashion dataset using our trained batch-hard triplet model. Our model is able to successfully retrieve photos across domains, that is, it can find the exact right product in a catalog of photos even when the query comes from a natural scene-image of lower quality. In many cases, it is also robust to pose changes and oblique capture angle differences between query and catalog photos. We do observe that the model performs better in retrieval for some categories more so than others. This is highly apparent in categories such as ``jeans", where the visual details of the intra-class examples are not as visually distinguishable as they are with ``blouses".

\subsection{Stanford Online Products dataset}
\begin{figure}
    \centering
    \includegraphics[width=0.45\textwidth]{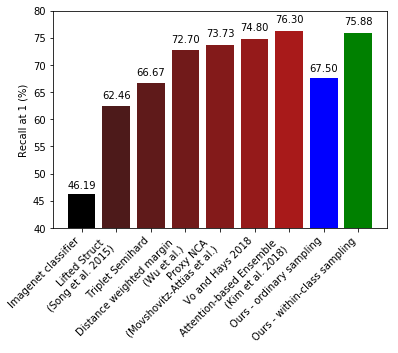}
    \caption{Recall at 1 on Stanford Online Products for our model and \cite{Song2016,wu2017sampling,Vo2018,Movshovitz-Attias2017,kim2018attention}.}
    \label{fig:sop_recall@1}
\end{figure}

We also evaluated our methods on the Stanford Online Product (SOP) dataset~\cite{Song2016}, which contains 120,053 images of 22,634 items from eBay, roughly equally distributed among 12 categories such as ``stapler" and ``bicycle." Unlike DeepFashion, SOP is not divided into separate domains for query and retrieval images; rather, the model should retrieve other images of the same product from the same set as the query was drawn from. Figure \ref{fig:sop_recall@1} shows results for our model and several other published results on SOP. Most previous work on this dataset reports only the recall at 1, so we compare based on this metric as well.

The first result with ``ordinary sampling" uses exactly the same model and training procedure as our best-performing model on DeepFashion. Although the datasets differ in that DeepFashion has separate ``consumer" and ``shop" domains, we showed that our model actually improves when ignoring these at training time. Therefore it is not surprising that the model also works well on SOP. However, the model trained with our ordinary sampling method significantly underperforms the state of the art on this dataset.

The second result with ``within-class sampling" uses the same model except that the minibatches are constructed differently to improve the triplet sampling. Some fraction of minibatches -- 80\% for the results shown here -- are constructed entirely of images from one of the SOP categories (``toaster," ``chair," etc.). The other minibatches are constructed as described in section \ref{sec:training}, to ensure that the model does not confuse images from different classes.

\begin{figure}
    \centering
    \includegraphics[width=0.45\textwidth]{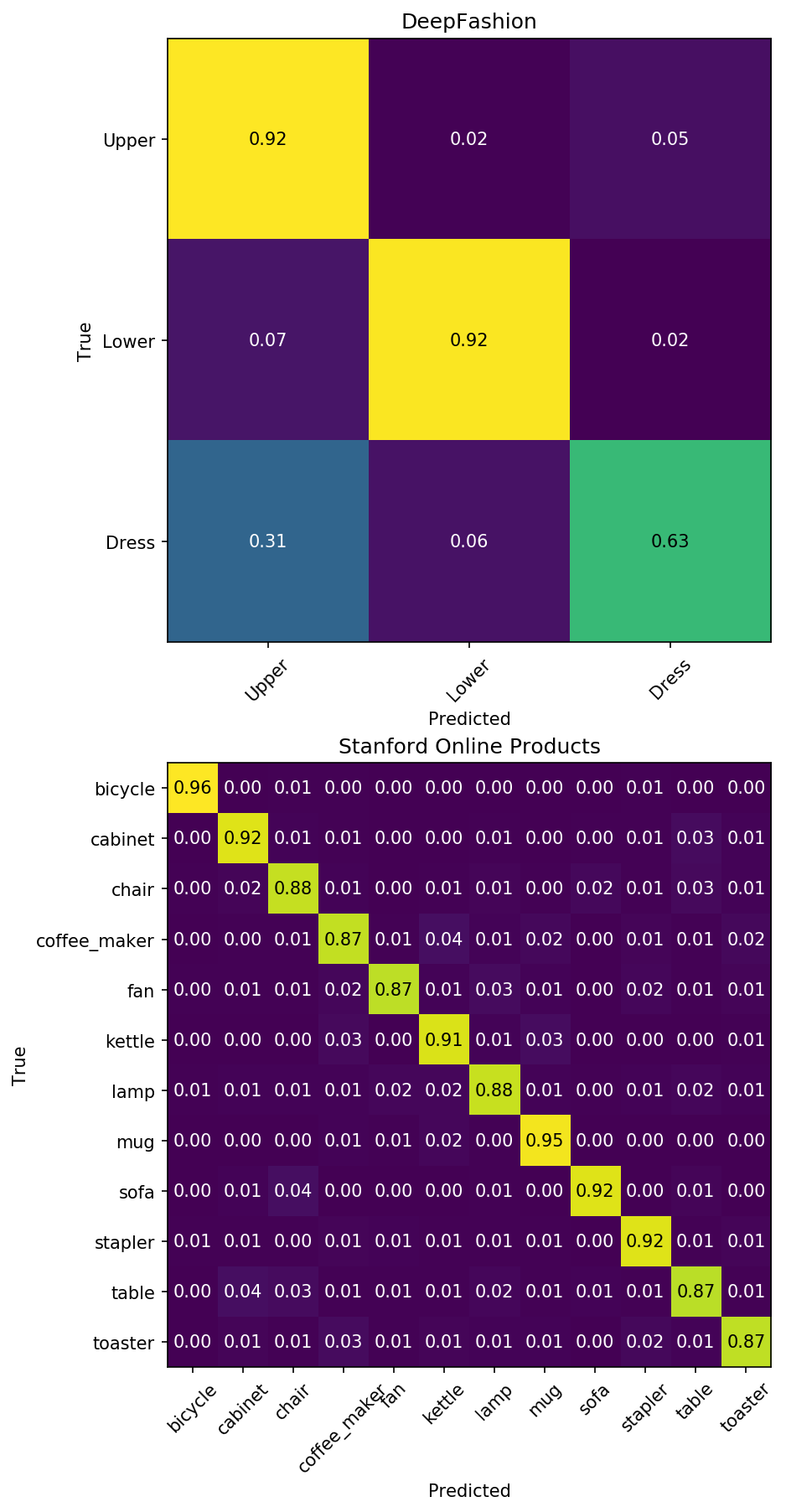}
    \caption{Class confusion matrices. The predicted class is the class of the first retrieved example.}
    \label{fig:confusion}
\end{figure}

Figure \ref{fig:confusion} illustrates why this within-class sampling improves the performance of our model on Stanford Online Products, while we found it had no benefit or even degraded performance on DeepFashion. The image categories in SOP are quite distinct, and the model rarely retrieves images from the wrong class as the first predicted result. For the confusion matrices in figure \ref{fig:confusion}, we treated the model as a classifier where the predicted class is the class of the first retrieved image. Even in the classes that generate the most confusion, an image is still matched with an image of the same class first 87\% of the time. In contrast, the model trained on DeepFashion makes mistakes outside the item's category as much as 31\% of the time, in the case of dresses. Note that we constructed the three DeepFashion categories in figure \ref{fig:confusion} by combining classes from among the 23 provided by the metadata whenever the distinction was unclear to us; evaluating classification accuracy in the same way on the original classes gives 59.8\% overall.

\begin{figure}
    \centering
    \includegraphics[width=0.45\textwidth]{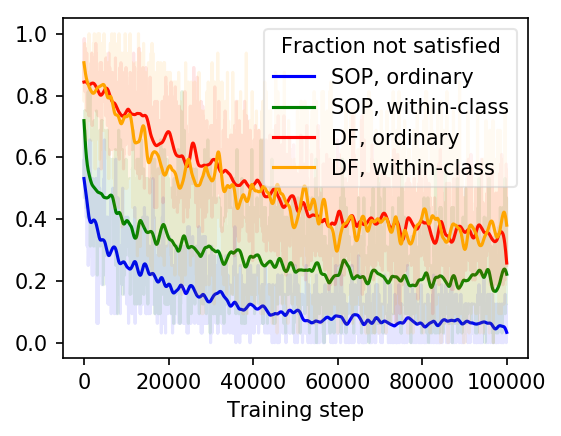}
    \caption{Effect of our ordinary vs mostly-in-class sampling on the fraction of batch-hard triplets giving nonzero triplet loss plotted for the DeepFashion (DF) and Stanford Online Product (SOP) datasets.}
    \label{fig:within-class}
\end{figure}

The high classification accuracy in the case of Stanford Online Products suggests that we can greatly increase the efficiency of finding triplets that contribute to the loss by forming minibatches from within the categories. Figure \ref{fig:within-class} shows that this technique is effective on SOP, increasing the fraction of batch-hard triplets with nonzero loss by as much as 20 percentage points. No such effect is observed in the case of DeepFashion, but the fraction of non-satisfied triplets is still larger on DeepFashion.

\subsection{Effect of batch size}

\begin{figure}
    \centering
    \includegraphics[width=0.45\textwidth]{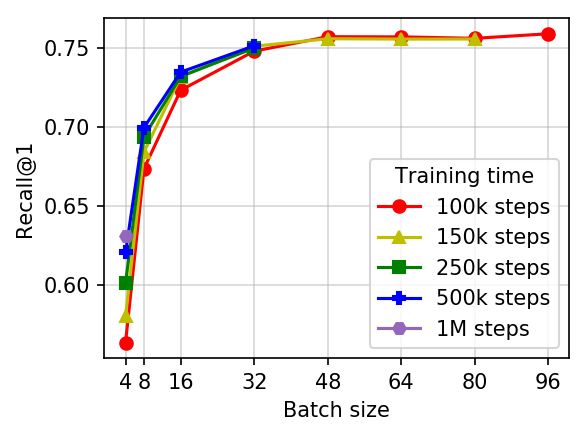}
    \caption{Small batch sizes lead to poor results, but the benefit of larger batches saturates around batch size 48. Results for Stanford Online Products shown; results for DeepFashion are similar.}
    \label{fig:batch_size}
\end{figure}

It is important to use sufficiently large minibatches to allow our method to find enough useful triplets to accumulate a meaningful signal for the gradient. We found that very small batch sizes led to poor performance, as shown in Figure \ref{fig:batch_size} for models trained on Stanford Online Products with 80\% of batches sampled from within a class. The benefit of increasing the batch size diminishes after batches of about 32 anchor-positive pairs, with no significant improvement seen for batch sizes larger than 48 pairs. We observed similar results on the DeepFashion dataset.

\section{Discussion}
We attribute the strength of our results on the DeepFashion Consumer-to-Shop Clothes Retrieval dataset compared to prior work to the effectiveness of our sampling approach. While random triplet sampling appears to fail because most triplets are do not provide a useful contribution to the gradient, the offline hard negative mining used in the previous state-of-the-art on DeepFashion~\cite{Gajic2018} may be limited for the opposite reason. That is, selecting too strongly for hard negatives may overemphasize mislabeled negatives and others that are so similar to the anchor that any differences the network finds may fail to generalize, leading to noisier gradients or a large train/test gap.

We further improved our results by ignoring the distinction between the ``consumer" and ``shop" domains in DeepFashion at training time.  We conjecture that this performance gain arises simply from increasing the effective size of the dataset by including single-domain pairs, and that the two domains in DeepFashion are not sufficiently distinct that the model would benefit much from specializing to cross-domain comparisons as was done in \cite{Gajic2018}.

The Recall-at-k results on the Stanford Online Products dataset are generally greater for a given model than on DeepFashion. We found that a more strongly selective sampling strategy, finding hard triplets from within single-class batches, improved results on SOP but not on DeepFashion. These results suggest that extra sampling strategies beyond batch-hard sampling may become more useful as the dataset gets ``easier" in the sense that a greater portion of the triplet constraints are easily satisfied. Our within-class sampling strategy in particular is likely to be useful whenever a model easily distinguishes the classes. In this way, additional labeling beyond the item identities may be leveraged indirectly to improve the training of a visual search model.

Although our results on Stanford Online Products are not clearly better than the state-of-the-art, our method is simple and general and achieves similar performance to Proxy NCA~\cite{Movshovitz-Attias2017} and other triplet methods such as \cite{Vo2018}. The latter paper focused on finding the optimal layer of the network to use as embeddings at test time, which may be considered independently of the sampling strategy we propose. As far as we know, the only published result that surpasses our recall-at-1 performance on Stanford Online Products used an ensemble method that could also be combined with our sampling strategies.

\section{Conclusion}
We have shown that a simple model based on a standard deep convolutional neural network and triplet loss, when trained with an effective online sampling technique, performs well on a visual search task. Our basic method surpasses the previous state of the art on the DeepFashion Consumer-to-Shop Clothes Retrieval dataset, and we observe that allowing consumer-consumer and shop-shop matches at training time improves performance further. We also evaluated on the Stanford Online Products dataset, and we achieved results similar to the state of the art after modifying our sampling strategy to favor minibatches from within a product category. Interestingly, this modified sampling strategy did not provide a similar benefit on DeepFashion. While our basic method, based on batch-hard sampling of triplets, appears to generalize well, specific adjustments may be appropriate on some datasets and not others. Future work may also combine the insights of this work with those of previous work to improve the state of the art for product visual search.
\newpage

\bibliographystyle{aaai}
\bibliography{retrieval_bib}

\begin{thebibliography}{}

\bibitem[\protect\citeauthoryear{Babenko \bgroup et al\mbox.\egroup
  }{2014}]{Babenko2014}
Babenko, A.; Slesarev, A.; Chigorin, A.; and Lempitsky, V.~S.
\newblock 2014.
\newblock Neural codes for image retrieval.
\newblock {\em CoRR} abs/1404.1777.

\bibitem[\protect\citeauthoryear{Chopra, Hadsell, and
  LeCun}{2005}]{chopra2005learning}
Chopra, S.; Hadsell, R.; and LeCun, Y.
\newblock 2005.
\newblock Learning a similarity metric discriminatively, with application to
  face verification.
\newblock In {\em Computer Vision and Pattern Recognition, 2005. CVPR 2005.
  IEEE Computer Society Conference on}, volume~1,  539--546.
\newblock IEEE.

\bibitem[\protect\citeauthoryear{Gajic and Baldrich}{2018}]{Gajic2018}
Gajic, B., and Baldrich, R.
\newblock 2018.
\newblock {Cross-domain fashion image retrieval}.
\newblock In {\em The IEEE Conference on Computer Vision and Pattern
  Recognition (CVPR) Workshops}.

\bibitem[\protect\citeauthoryear{He \bgroup et al\mbox.\egroup }{2016}]{He2016}
He, K.; Zhang, X.; Ren, S.; and Sun, J.
\newblock 2016.
\newblock {Identity Mappings in Deep Residual Networks}.
\newblock  1--15.

\bibitem[\protect\citeauthoryear{Hermans, Beyer, and Leibe}{2017}]{Hermans2017}
Hermans, A.; Beyer, L.; and Leibe, B.
\newblock 2017.
\newblock {In Defense of the Triplet Loss for Person Re-Identification}.

\bibitem[\protect\citeauthoryear{Huang \bgroup et al\mbox.\egroup
  }{2015}]{Huang2015}
Huang, J.; Feris, R.; Chen, Q.; and Yan, S.
\newblock 2015.
\newblock {Cross-domain image retrieval with a dual attribute-aware ranking
  network}.
\newblock {\em Proceedings of the IEEE International Conference on Computer
  Vision} 2015 Inter:1062--1070.

\bibitem[\protect\citeauthoryear{Kiapour \bgroup et al\mbox.\egroup
  }{2015}]{Kiapour2015}
Kiapour, M.~H.; Han, X.; Lazebnik, S.; Berg, A.~C.; and Berg, T.~L.
\newblock 2015.
\newblock {Where to buy it: Matching street clothing photos in online shops}.
\newblock {\em Proceedings of the IEEE International Conference on Computer
  Vision} 2015 Inter:3343--3351.

\bibitem[\protect\citeauthoryear{Kim \bgroup et al\mbox.\egroup
  }{2018}]{kim2018attention}
Kim, W.; Goyal, B.; Chawla, K.; Lee, J.; and Kwon, K.
\newblock 2018.
\newblock Attention-based ensemble for deep metric learning.
\newblock {\em arXiv preprint arXiv:1804.00382}.

\bibitem[\protect\citeauthoryear{Liu \bgroup et al\mbox.\egroup
  }{2016}]{Liu2016}
Liu, Z.; Luo, P.; Qiu, S.; Wang, X.; and Tang, X.
\newblock 2016.
\newblock {DeepFashion: Powering Robust Clothes Recognition and Retrieval with
  Rich Annotations}.
\newblock {\em 2016 IEEE Conference on Computer Vision and Pattern Recognition
  (CVPR)} (1):1096--1104.

\bibitem[\protect\citeauthoryear{Movshovitz-Attias \bgroup et al\mbox.\egroup
  }{2017}]{Movshovitz-Attias2017}
Movshovitz-Attias, Y.; Toshev, A.; Leung, T.~K.; Ioffe, S.; and Singh, S.
\newblock 2017.
\newblock {No Fuss Distance Metric Learning using Proxies}.
\newblock {\em Iccv}  360--368.

\bibitem[\protect\citeauthoryear{Russakovsky \bgroup et al\mbox.\egroup
  }{2015}]{ILSVRC15}
Russakovsky, O.; Deng, J.; Su, H.; Krause, J.; Satheesh, S.; Ma, S.; Huang, Z.;
  Karpathy, A.; Khosla, A.; Bernstein, M.; Berg, A.~C.; and Fei-Fei, L.
\newblock 2015.
\newblock {ImageNet Large Scale Visual Recognition Challenge}.
\newblock {\em International Journal of Computer Vision (IJCV)}
  115(3):211--252.

\bibitem[\protect\citeauthoryear{Schroff, Kalenichenko, and
  Philbin}{2015}]{Schroff2015}
Schroff, F.; Kalenichenko, D.; and Philbin, J.
\newblock 2015.
\newblock {FaceNet: A unified embedding for face recognition and clustering}.
\newblock {\em Proceedings of the IEEE Computer Society Conference on Computer
  Vision and Pattern Recognition} 07-12-June:815--823.

\bibitem[\protect\citeauthoryear{Song \bgroup et al\mbox.\egroup
  }{2016}]{Song2016}
Song, H.~O.; Xiang, Y.; Jegelka, S.; and Savarese, S.
\newblock 2016.
\newblock Deep metric learning via lifted structured feature embedding.
\newblock In {\em IEEE Conference on Computer Vision and Pattern Recognition
  (CVPR)}.

\bibitem[\protect\citeauthoryear{Vo and Hays}{2018}]{Vo2018}
Vo, N., and Hays, J.
\newblock 2018.
\newblock {Generalization in Metric Learning: Should the Embedding Layer be the
  Embedding Layer?}

\bibitem[\protect\citeauthoryear{Wu \bgroup et al\mbox.\egroup
  }{2017}]{wu2017sampling}
Wu, C.-Y.; Manmatha, R.; Smola, A.~J.; and Kr{\"a}henb{\"u}hl, P.
\newblock 2017.
\newblock Sampling matters in deep embedding learning.
\newblock In {\em Proc. IEEE International Conference on Computer Vision
  (ICCV)}.

\end{thebibliography}

\end{document}